\documentclass[letterpaper, 10 pt, conference]{ieeeconf}  
\pdfoutput=1

\IEEEoverridecommandlockouts                              

\usepackage[utf8]{inputenc}
\usepackage{mathtools}
\usepackage{amsmath}
 \numberwithin{equation}{subsection}
 \usepackage{graphicx}
 \usepackage{amsmath}
 \usepackage{booktabs}
 \usepackage{tabularx}
 \usepackage{graphicx} 
\usepackage{caption}
\usepackage[table]{xcolor}
\usepackage{algorithm}
\usepackage{algpseudocode}
\usepackage{placeins}
\makeatletter
\def\BState{\State\hskip-\ALG@thistlm}
\makeatother
\title{\LARGE \bf Investigating  Resistance of Deep Learning-based  IDS against Adversaries using  min-max Optimization
}

\author{ \parbox{4 in}{\centering Rana AbouKhamis, Omair Shafiq, Ashraf Matrawy
       \thanks{}\\
       School of Information Technology\\
       Carleton University\\ Ottawa, Ontario, Canada.\\
       {\tt\small rana.aboukhamis@carleton.ca}
         {\tt\small omair.shafiq@carleton.ca},
        {\tt\small ashraf.matrawy@carleton.ca}
        }
}

\begin{document}

\maketitle
\thispagestyle{empty}
\pagestyle{empty}

\begin{abstract}
With the growth of adversarial attacks against
machine learning models, several concerns have emerged about
potential vulnerabilities in designing deep neural network-based
intrusion detection systems (IDS). In this paper, we study the
resilience of deep learning-based intrusion detection systems
against adversarial attacks. We apply the min-max (or saddle-point)
approach to train intrusion detection systems against
adversarial attack samples in NSW-NB 15 dataset. We have the
max approach for generating adversarial samples that achieves
maximum loss and attack deep neural networks. On the other
side, we utilize the existing min approach [2] [9] as a defense
strategy to optimize intrusion detection systems that minimize
the loss of the incorporated adversarial samples during the
adversarial training. We study and measure the effectiveness
of the adversarial attack methods as well as the resistance
of the adversarially trained models against such attacks. We
find that the adversarial attack methods that were designed in
binary domains can be used in continuous domains and exhibit
different misclassification levels. We finally show that principal
component analysis (PCA) based feature reduction can boost
the robustness in intrusion detection system (IDS) using a deep
neural network (DNN).

\end{abstract}

\textit{Keywords: Deep Learning-based Intrusion Detection, adversarial samples, adversarial learning.
}

\section{Introduction}

The Security applications of deep neural networks (DNNs) like Intrusion Detection System (IDS), malware detection, spam-filtering have become essentials in designing tasks for data protection, classification, and prediction. These different type of tasks are relying on the intelligence to build a model that typically classify and discriminate between "benign" and "malign" samples, like attack and benign packets. With the rapid increase of using DNNs and the vulnerability of DNNs to adversarial attacks, the sophistication of attack techniques tools is also increased.
Therefore, various researches \cite{wang2018deep}\cite{homoliak2018improving} find that different attacks add severe challenges to vulnerabilities of DNN architecture design. The fact that the training of DNNs is based on data, the classification task can be manipulated by crafted, and perturbation inputs called adversarial samples. Adversaries samples are often visually imperceptible, and they are designed to reliably mislead a machine learning model toward incorrect classification and evade detection \cite{szegedy2013intriguing}.
Toward covering the issue of securing classifiers against adversarial attacks, we study the adversarial attacks against DNNs and their robustness. In this paper, our primary goal is to develop a deep learning-based IDS defender model that can demonstrate robustness against different adversarial attacks that prove their effectiveness in different classification domains, including image and malware classifiers.

In this paper, we focus on the following:

\begin{enumerate}

\item  Utilizing existing methods that use saddle-point formulation and adversarial training as a defense strategy to optimize an existing intrusion detection system framework [2] that can robustly handle adversarial attacks and reduce false negative (FN) rates and thus increase model robustness. We train the optimized network models using NSW-NB 15 data.

\item Analyzing four existing methods \cite{al2018adversarial} to generate adversarial samples including Projected Gradient Descent (PGD) and Fast Gradient Simple Method (FGSM) across DNNs. We aim to investigate if different adversarial attacks including Multi-Step Bit Gradient Ascent ($BGA^S$) and Bit Coordinate Ascent ($BCA^S$) \cite{al2018adversarial} which are designed for discrete domain can generate adversarial samples in continuous domain. We observe that BGA adversarial samples have the highest evasion rate among all other adversaries.

\item Conducting two sets of experiments with different approaches in pre-processing NSW-NB 15 dataset and structuring the DNNs detection models. We apply Principal Component Analysis (PCA) based dimensionality reduction on the NSW-NB 15 dataset to attempt lower evasion attack rates.

\item We select DNNs to develop our IDS model, experimental models, using Pytorch \cite{al2018adversarial}. We also compare several existing machine learning algorithms including Neural Network (NN), Random forest (RF), AdaBoost, Naive Bayesian (NB), and SVM for classifying network traffic. The experiment result shows that NN has a reasonable detection accuracy compared to the other approaches. 
\end{enumerate}

We structure the remainder of the paper as follows. In Section II, we begin a brief background about neural networks and deep learning, followed by a detailed overview of adversarial attack methods and adversarial deep learning. In Section III, we survey some related work. In Section IV, we explorer our experiments and methodology, starting by describing the dataset and how we preprocess it following with our IDS prototype. In section V, we include the IDS algorithm and the actual experiments. In Section VI, we present and evaluate our experimental results started by our evaluation metrics and ended by our discussion. Finally, in Section VII, we summarize and present some recommendations on how we can strengthen our work in future research.

\section{Background}

An intrusion detection system (IDS) is a key component in network systems to monitor and analyze real-time network activities for any symptoms of suspicious, anomalous activities and issue alerts when such types of activities are discovered. One of the significant limitations of the standard intrusion detection systems (IDS) is the urgency to filter and reduce false alarms\cite{homoliak2018improving}. Toward this direction, a majority of IDSs enhance their capabilities by using neural networks (NN) towards deep learning. As a solution, Deep Neural Networks (DNNs) based IDS solutions have been developed to make better learning and processing for big data and variety of attacks for future prediction.

\subsection{Neural Networks and Deep Learning}
Most deep learning models are structured based on Neural Networks with multiple hidden layers. This called Deep Neural Networks (DNNs). DNN efficiency is related to dataset size and requires massive computing power. Using GPUs for training with DNN models accelerate the learning process. The DNN architecture is made by many neurons that are connected to other neurons. Each neuron connection is associated with weight to multiply it with input variables (feature) by the non-linear activation function $\varphi$($x$). The activation function maps the output of neuron values depending on the function. In our paper, the activation function $\varphi$ for the output is LogSoftMax and ReLU for the three hidden layers. To compare how far the prediction result from the target value $y$ $\in$ $\mathcal{Y}$ and to achieve an optimal model, we use a loss (or cost) function $ \mathcal{T}(\emptyset,x,y)$ \cite{wang2018deep}.
The network neurons keep adjusting the values of their parameters until achieving close prediction to reduce error \cite{al2018adversarial}. 
\subsection{Adversarial Deep Learning}
It is increasingly important to guarantee the protection and robustness against adversarial manipulation. 
Adversaries can evade classifiers by adding a calculated perturbation $\gamma$ to legitimate samples $x \in \mathcal{Z}$(x) to create a new version $x^* \in \mathcal{Z^*}(x)$ called "Adversarial Sample" , where $\mathcal{Z^*}(x) \subseteq  \mathcal{Z}(x)$, $\mathcal{Z}$ is a set of allowed perturbations and $\mathcal{Z^*}$ is a set of adversarial samples that takes into consideration the max norm perturbations. 
In our work with continuous space, we focus on $\mathcal{L}_\infty$\ to gives the max value of the parameters among each element samples: $||x||_\infty$= $max_i |x_i|$. 
\subsubsection{Adversarial Learning}
\hfill\\
To increase trained model robustness against adversarial samples, we use the adversarial learning defender strategy against adversarial attacks that incorporating adversarial samples in the training phase. Our learned IDS model can be manipulated by adversaries $\gamma$ that maximize the loss $\mathcal{T}$ to classify attack samples as benign samples.

We combined in (II-B.1) \cite{al2018adversarial} between the adversarial crafted formulation (the inner maximization problem); that aim to find an adversarial sample $x^*$ from original sample $x$ that achieves a high loss; and the defender DNN model (the outer minimization problem ); that aim to minimize the adversarial loss and increase model robustness. This combination is the use of \textbf{min-max (or saddle-point)  formulation} \cite{madry2017towards}\cite{al2018adversarial}. The use of this formulation allows us to achieve high robustness against adversarial attacks.
\begin{equation}
\emptyset^* \underbrace{ \in  
 \underbrace{{\operatorname{arg}\underset{\emptyset \in \Theta}{\operatorname{min}} \mathcal{T}(\emptyset,x,y)}}_\text{outer minimization }
[\overbrace{\underset{x^* \in \mathcal{Z}}{\operatorname{max}} \mathcal{T}(\emptyset,x^*,y)}^\text{inner maximization }]}_\text{adv learning}
\end{equation}

Strategies for creating and designing adversarial attacks are affected by different limitations. The capability of attacker relays on how attacker controls the input data and the ability in manipulating the data \cite{biggio2013security}\cite{biggio2018wild}. Following section focuses on adversarial attack methods and how to generate adversarial samples.

\subsection{Adversaries Attack Methods}
\subsubsection{\textbf{Fast Gradient Sign Method (FGSM)}}
It is about updating the gradient of the loss function \begin{math}\mathcal{T}\end{math} along with its sign direction in one step. The  $\nabla$ is used to calculate the derivative of the loss function. The derivative value of \begin{math} T (\emptyset,x ,y)\end{math} is used to find the slope of the function at a given point to adjust the values of the parameters whether by increasing or decreasing them. This adjustment is a small value $\rho$ that makes the perturbation not distinguishable.

\subsubsection{\textbf{$FGSM^S$ with Multiple Step} \cite{kurakin2016adversarial}}
It is a way to apply Fast gradient sign method (FGSM) multiple times with a small step size.

Prior works \cite{al2018adversarial} \cite{hu2017generating} make use of two approaches in gradient-based methods (e.g update weights in DNN and crafting adversarial samples):

\emph{Randomize Rounding Approach ($rFGSM^S$)} is used to find optimal solution in fraction problem. It’s called linear programming randomize rounding that gives precision integer result on each round. 
 \emph{Deterministic Approach ($dFGSM^S$) } can be used for both continuous and discrete linear optimization problems \cite{rader2010deterministic} make the outcome deterministic. This approach depends on the attacker level of knowledge about the classifier \cite{jain2004development}. 
In this work, we use $dFGSM^S$ that uses Deterministic approach and $rFGSM^S$ that uses randomize rounding approach to craft our adversarial samples $x^*$.
\subsubsection{\textbf{Multi-Step Bit Gradient Ascent ($BGA^S$)}}
In the continuous domain, the use of multiple iterations like $dFGSM^S$ and $rFGSM^S$ can generate only a single adversarial sample. We use the gradient to generate multiple adversarial samples and select the sample that gives the maximum loss using $\ell_2$ norm. 
In our work, we apply this method to our continuous features, and we use the $\ell_2$ norm to find the shortest distance from one point to another. 
\subsubsection{\textbf{Bit Coordinate Ascent ($BCA^S$)}}
In \cite{al2018adversarial}, the authors used this attack approach to update one bit each iteration that has a higher partial derivative of the loss. We use a similar approach to update our one continuous feature a time that achieve higher partial derivative of the loss compared to the other features.

\section{LITERATURE REVIEW}

Researchers perform various studies of adversarial attacks with their countermeasures and how they can easily deceive machine learning systems. 
In \cite{al2018adversarial}, authors presented different adversarial attacks methods to generate an adversarial example of binary malware file that preserves its functionality. They present a framework for training robust malware detection models by utilizing the saddle-point formulation that consists of the inner maximization and outer minimization problems. Their experiment result shows that rFGSM adversarial trained model has the lowest evasion rate and higher accuracy among all adversarial attacks. They used the Portable Executable (PE) files as a dataset created from VirusShare. BCA and BGA attacks have relatively low evasion rates comparing to the FGSM attacks. However, the BCA model has a high evasion rate among all adversaries attack, which indicates the lowest robustness comparing to other models.

In Mardy et al. \cite {madry2017towards} research, they studied the adversarial robustness of neural networks to guarantee security and be resistant to a broad class of attacks by using min-max (or saddle-point) formulation. Their study allows them to cast both attack and defense. 
They trained networks on MNIST and CIFAR10 models based on the saddle-point formulation and uses their optimal first-order adversary to be robust to a broad range of adversarial attacks. MNIST model accuracy against adversarial attack achieves 89\%, and CIFAR10 model achieves 46\% accuracy of the same attacks. They found that CIFAR10 has low accuracy.

While in \cite{wang2018deep}, the authors studied the vulnerabilities of deep learning-based IDS among adversarial attacks by evaluating the effectiveness of state-of-the-art attack algorithms using NSL-KDD dataset. They compare different adversarial attacks. Their experiments result shows that CW attacks have the lowest effectiveness. Also, the use of adversarial samples targeting specific model can fool another model. This adversarial property called transferability.

There are fewer works about utilizing min-max formulation and deep learning techniques to design optimal intrusion detection system that can robustly handle different types of adversarial attacks. In this paper, we built an experimental prototype and testbed based on the existing techniques \cite{al2018adversarial} \cite{madry2017towards} that
incorporates crafted adversarial samples during the training process and successfully achieve
robustness against the adversarial attack methods.
\section{Methodology}
\subsection{Dataset and Data Preprocessing}
All experiments described in this work are performed using UNSW-NB 15 dataset \cite{moustafa2015unsw}. UNSW-NB 15 dataset is a raw network flow that is created by the IXIA perfecStorm tool in a lab that generate normal activities and attack behaviors. 
Each row in the dataset has 49 features including the class label.
In our experiments, we use 82,332 records and split it into "Attack set" and Benign set" including 45,332 attack samples and 37,000 benign samples. 
The Attack set has 27,198 training samples and 181,34 samples between test and validation samples. For the Benign set, it has 22,200 training samples and 14,800 samples between test and validation samples.

For preprocessing UNSW-NB 15 dataset, we use the open source project Orange Data Mining version 3.20 \cite{JMLR:demsar13a} with its widgets to preprocess the dataset. We select \emph{\textbf{"Feature Selection"}} to assess the usefulness of our features. Based on the feature rank (or score), we select the best 28 features.
Also, we apply \emph{\textbf{"Normalization"}} to change the values of the numeric features that have a different range to increase classifier performance and accuracy. Moreover, we preprocess the dataset by using Principal Component Analysis (PCA) \cite{oja1992principal} to reduce the high dimension of the dataset while making sure not to lose the necessary information during reduction and keep variation. We use PCA in the second set of experiments.  

\begin{algorithm}[t]
\label{algo}
\caption{Experimental process based on \cite{al2018adversarial}}
\begin{algorithmic}[1]
\State\textbf{Input: }{ Attack and Benign training set from $\mathcal{D}$}
\State \textbf{Output: }{Adversarial trained model, $x^*$ adversaries}
\State Load attack and benign data (train, test \& validation)
\State Extract features  $x$ = \big\{x1,x2,...,xn \big\}
\State Construct DNN model $\mathcal{C}$
\State Define loss function $\mathcal{T}$ using "ADAM"
\State Define inner-maximization M
\State $Batch \leftarrow 100$
\Repeat 
\State Read $Batch$ of samples 
\If {Evasion method \textbf{!=} \emph{Natural}}
\State $Batch^* \leftarrow M(Attack\_Batch,T,Evasion)$ 
\State start Adversarial Learning ($Batch^*$)
\State do Test($Batch^*$)
\Else { Evasion method == \emph{Natural} }
\State start train ($Batch$)
\State do Test ($Batch$)
\EndIf
\Until{epoch=100 and C network converge}
\Procedure{inner-maximization}{x,y,T,s,method}
  \State Computer natural loss for original samples
  \State Initialize starting point
  \State Compute natural loss for original samples
  \State Compute gradient for the loss T
  \State Compute the new adversarial sample
  
\EndProcedure
\end{algorithmic}
\end{algorithm}

\subsection{Architecture Characteristics and Learning Setup}

The DNNs architecture is constructed with three hidden layers of 300 neurons in layer 1, 100 neurons in layer 2, and 40 neurons for layer 3. We applied two activation functions, ReLU \cite{agarap2018deep}, to the three hidden layers and LogSoftMax to the output layer to respond to the benign and attack output labels. We use the open-source machine learning library PyTorch to implement our five models, and we performed our experiments using the free Jupyter notebook cloud environment called 'Colaboratory' with GPU acceleration \cite{cola}.

The optimized model classifier $\mathcal{C}$ is trained using 45,332 attack samples and 37,000 benign samples. We use \emph{ADAM} as an optimizer algorithm to adjust the classifier $\mathcal{C}$ parameters $\emptyset^*$ based on (II-B.1).
The DNNs training without evasion attack achieves 94.1\% accuracy on the "benign dataset" and 81.6\% accuracy for the "attack dataset." The inner maximization was set up to run s=50 iterations.

\begin{figure*}[t]
\includegraphics[width=\textwidth,height=9cm]{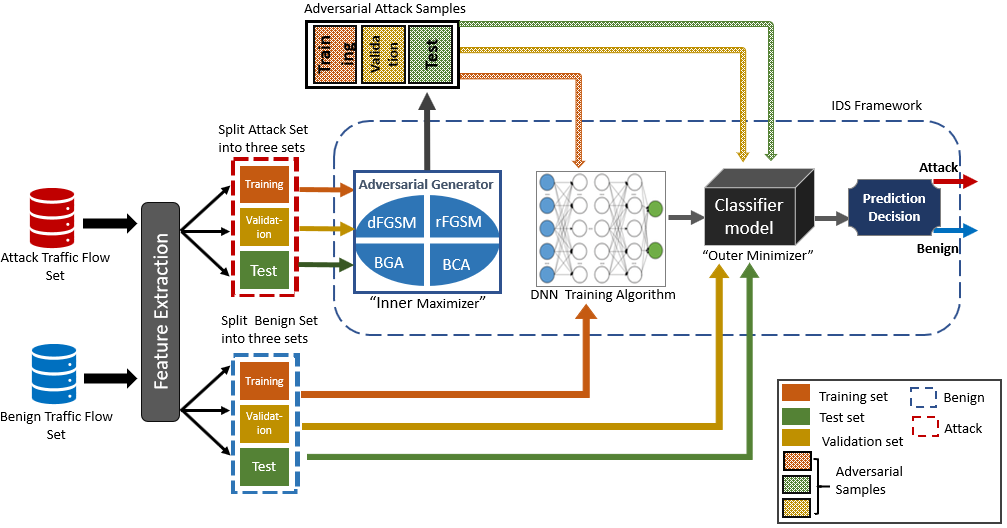}

  \caption{IDS prototype Architecture with the inner-maximizer that generate adversarial samples and the outer-minimizer that reduce False Negative (FN) rates and increase robustness}
  \label{fig:dnn}
  \centering
\end{figure*}

\section {Experiments}
In our paper, we assume our adversarial attacks are white-box evasion attacks. They attack the optimized IDS model during the test time to misclassify the positive "attack" sample as a negative "benign" sample.

We design \textbf{\emph{Algorithm I}} based on the existing works \cite{al2018adversarial} to craft adversarial samples using the "inner maximization" that generate adversarial samples and maximize the loss rate $\mathcal{T}$. The deep learning-based IDS defender solution is illustrated in Figure \ref{fig:dnn}, where the attack set and benign set are split into three sets (training, validation, and test). Also it shows how the attack set passes the inner maximizer to generate adversarial samples and then incorporate them in the learning phase as well. 
We train four models using the crafted adversarial samples generated by dFGSM, rFGSM,BGA, and BCA attack methods using the "inner maximization." We inject the adversarial samples into the training dataset to increase the model robustness.

We conducted two sets of experiments to evaluate the effectiveness of the four evasion attacks and the robustness of the IDS platform. The details of each experiment are described below.

\textbf{Algorithm Selection:} To do so, we first built five models (or binary classifiers) with different machine learning algorithms as shown in Table \ref{models} including Neural Network (NN) and trained them by our preprocessed and clean "UNSW-NB 15" dataset. The result of the experiments shows that NN has a reasonable detection of 92\% accuracy, 92\% Recall (or FPR), and also 92\% Precision (or FNR) to develop our deep learning-based IDS. 

\begin{table}[b]
\begin{tiny}
\caption{Performance statistics for UNSW-NB 15 dataset using different machine learning algorithms }
\label{models}
\resizebox{\columnwidth}{!}{

\begin{tabular}{ |c|c|c|c|c| } 
\hline
Method & Accuracy &  AUC & Precision & Recall\\ 
\hline
NN & 92 & 98 & 92 & 92\\
\hline
 Random Forest & 94 & 98 & 94 & 94\\
\hline
AdaBoost & 94 & 98 & 94 & 94\\
\hline
Naive Bayes & 73 & 84 & 75 & 73\\
\hline
Constant & 55 & 50 & 30 & 55\\
\hline
SVM & 45 & 46 & 46 & 45\\
\hline
\end{tabular}
}
\end{tiny}
\end{table}

\subsection{Experiments I}
In the first set of experiments I, our objective is to identify the most robust training model to adversarial samples for our IDS prototype. Thus, we extracted 28 attributes (features) from our preprocessed dataset to train our IDS prototype. Then, we load the Attack and Benign sets after splitting them into training, validation, and test samples. In experiment I, we set up the hyper-parameters as following: batch size= 100 sample, learning rate $\rho$ = 0.01, evasion rate=50, and 100 epochs. We build five adversarial models including the Natural model. 
\subsubsection{Natural Model}
First, we build our first model naturally using the non-crafted Attack (clean) samples and non-crafted Benign samples. We achieve 94.1\% for benign accuracy and 81.6\% for attack accuracy with 87.85\% for overall accuracy. We also evaluate the robustness of the "Natural" model against the four adversarial evasion attacks.
\subsubsection{Adversarial Models}
Second, we trained four models using the outer minimization algorithms against all adversarial samples generated by the inner maximization. Afterward, we evaluate each model robustness against all adversarial attack samples and clean sets.

\begin{figure}{}
\includegraphics[width=\linewidth,keepaspectratio=true]{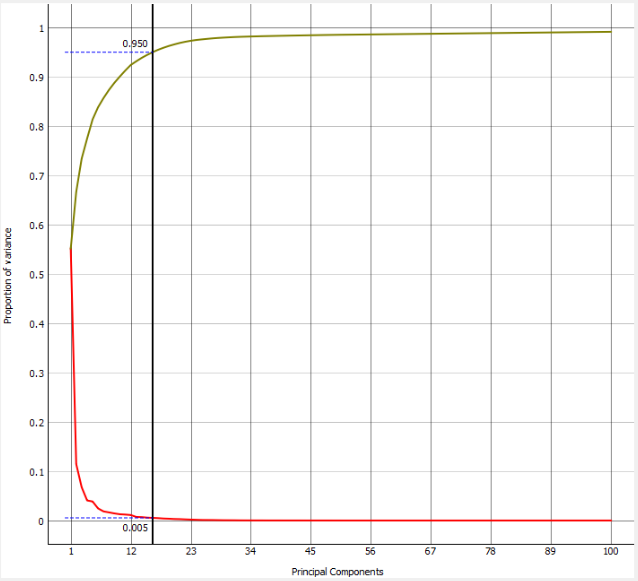}
  \caption{Principal Component Analysis (PCA) figure selecting 16 principal components with 95\% variance}
  \label{fig:PCA}
  \centering
\end{figure}
\subsection{Experiments II}
In the second set of experiments, we use the same UNSW-NB 15 dataset \cite{moustafa2015unsw}. However, this experiment starts by preprocessing and transforming the dataset differently using PCA before we load it to our DNNs.  Figure \ref{fig:PCA} shows that the 16 obtained principal components are covering 95\% variation. This introduced DNNs architecture still has three hidden layers, but all three layers have 200 neurons. The hyper-parameters of experiment II for learning rate, epoch size, batch size, and evasion rate are 0.001, 150, 8, and 50, respectively. 
The reason behind introducing a new optimization DNNs architecture using PCA is to investigate if applying PCA will reduce the evasion rate of the studied  adversarial  attacks and enhance the adversarial trained models robustness.
Therefore, we generate adversarial samples using the four adversarial attack methods generated by the inner maximizer. Also, we built five models, including a Natural model and the four adversarial training models that are trained with crafted adversarial samples.


\section{Results and Evaluation}

This section evaluates the IDS prototype that we implement using DNNs architecture. 

Before diving into the results and evaluation, we want to present the measurement metrics we use to evaluate the results.
We measure our experiment performance based on the classification Accuracy (AC), Evasion rate(ER), and Covering Number (CN). 
\emph{Evasion Rate or False Negative Ratio (FNR)} is the ratio of a total number of crafted samples that are classified as benign samples among total attack samples.
\emph{Covering Number (CN)} is a measurement \cite{al2018adversarial} to asses the inner maximizer algorithm's ability to create adversarial samples. CN is the ratio of the total number of adversarial attack samples generated among the original samples in the training epochs. Therefore, high CN indicates a high robustness model \cite{al2018adversarial}. The CN is computed and updated in each training epoch.


\begin{table}[t]
\begin{tiny}
\caption{ Evasion Rate Results of Experiment I. The model column indicates the trained models by the outer minimizer. The adversarial attack methods row indicates the methods used that generate adversarial samples by the inner maximizer}
\label{evasionRate}
\resizebox{\columnwidth}{!}{
\begin{tabular}{ |c|c|c|c|c| } 
\hline
- & \multicolumn{4}{|c|}{Adversarial Attack Methods}\\
\hline
Model &   dFGSM & rFGSM & BGA & BCA\\
\hline
Natural & 100 & 99.9 & 99.9 & 64.4\\
\hline

 dFGSM& 17.5 & 18 & \cellcolor{blue!25}\textbf{15.7}& \cellcolor{gray!25}\textbf{23}\\
\hline
rFGSM  & 23.3 & \cellcolor{blue!25}\textbf{19.1} &   \cellcolor{gray!25}\textbf{28.6} &   23.0 \\
\hline
BGA & \cellcolor{blue!25}\textbf{21.3} & \cellcolor{gray!25}\textbf{22.6} & 21.8 & \cellcolor{gray!25} \textbf{22.6}\\
\hline
BCA  & 21.5 & \cellcolor{blue!25}\textbf{18.9} & \cellcolor{gray!25}\textbf{24.2}& 21.6\\
\hline
\end{tabular}
}
\end{tiny}
\end{table}  
\begin{table}[h]
\begin{tiny}
\caption{Evaluation Metrics including Overall Accuracy, Evasion rate of models with regard to the corresponding adversary (FNR), FPR and the Cover Number (CN)) rate for Experiment I}
\label{metrics}
\resizebox{\columnwidth}{!}{

\begin{tabular}{ |c|c|c|c|c| } 
\hline
Model &  Accuracy &  FPR & Evasion Rate/FNR & CN \\ 
\hline

Natural &    87.2 &  5.9 & 18.4 & 1.0\\
\hline
dFGSM &      87.4 &  6.6 & \cellcolor{blue!25}\textbf{17.5} & 1.3\\
\hline
rFGSM &      87.1 &  5.4 & 19.1 & \cellcolor{yellow!25}\textbf{7.9}\\
\hline
BGA   &      86.3 &  3.7 & \cellcolor{brown!25}\textbf{21.8} & 2.7\\
\hline
BCA   &      86.4 &  3.9 & \cellcolor{brown!25}\textbf{21.6} & 1.2\\
\hline
\end{tabular}
}
\end{tiny}
\end{table}

\subsection{Results of Experiment I}
We evaluate the effectiveness of the adversarial samples and the robustness of the defender trained models against the crafted adversarial samples. Table \ref{evasionRate} summarizes the results of Experiment I for the five trained models, including the Natural model and the four adversaries attack methods. The result values in this Table \ref{evasionRate} are the evasion rate of the adversarial attacks using the perturbation attack set. We observe the following:
The evasion rates for the "Natural" training are high [64.4\% - 100\%] compared to the other adversarial training models. This result is expected because the "Natural" training model is not trained by any adversarial samples and only trained by a clean dataset. As we can see, the Natural training method results demonstrate the effectiveness of the evasion attacks that induce significant changes inaccuracy. The evasion rate reaches 100\% with 0\% accuracy against $dFGSM^s$ adversaries samples. We observe a decrease in the evasion rate with $BCA^s$ attack against the Natural training model from 100\% to 64.6\%. This shows that $BCA^s$ confidence is reduced compared to the other adversarial training models. The most robust models against the adversaries are shaded in blue in Table \ref{evasionRate}. While the ones shaded in gray are the most powerful attacks. We can see from Table \ref{evasionRate} that $dFGSM^s$ model has relatively the lowest evasion rates 17.5, 18, 15.7 and 23 for $dFGSM^s$,$rFGSM^s$, $BGA^s$ and $BCA^s$ respectively across all adversarial attacks especially against $BGA^s$. The $BGA^s$ evasion rate achieves the lowest 15.7 as shown in Table \ref{evasionRate}. However, we can see in Table \ref{evasionRate} that $BGA^s$ adversarial attack method outperforms other adversarial attack methods with evasion rate of 28\% against $rFGSM^s$ model. 

In general, we note as we expect that all training methods are relatively robust to adversaries from the same methods. However, we notice that some models are more robust to other adversaries. For example, $dFGSM^s$ model has 15.7\% evasion rate for $BGA^s$ while the evasion rate for same $dFGSM^s$ attack method is high 17.5\% comparing to 15.7\% evasion rate for $BGA^s$.   

We see in Table \ref{metrics} that $dFGSM^s$ has the lowest FNR=17.5\% comparing to $rFGSM^s$, $BGA^s$ and $BCA^s$. This result corresponds to our finding in Table \ref{evasionRate} that $dFGSM^s$ has the lowest evasion rate across all adversarial attack methods. While $BGA^s$ and $BGA^s$ models FNR rates are 21.8\% and 21.6\%, respectively. This considers the highest rate compared to the other models. 

Moreover, we computed the Covering Number (CN) rate for all training models, as shown in Table \ref{metrics} against the same attack method to measure the ratio of a total number of crafted samples to the total original attack samples in each epoch. The main objective of CN metric computation is to observe which adversarial attack method can cover more adversarial samples. Table \ref{metrics} in the yellow shaded cell shows that $rFGSM^s$ attack method has a higher CN ratio of 7.9. This high rate of CN means that the $rFGSM^s$  attack method explored almost eight times more attack samples compared to the CN of the Natural model with CN=1. Adversarial attack method with high CN explored more adversarial samples during training time, which leads to a more robust model.
In conclusion, we notice relatively that $dFGSM^s$ is the most robust model in experiment I and $BGA^s$ is the most successful evasion attack.


\subsection {Results of Experiment II using PCA}
In this experiment, we build a second DNN model with different architecture, and we preprocess the UNSW-NB 15 set using PCA. The objective is to improve the resilience of the IDS model to adversarial attacks.
As shown in Table \ref{evasionRatePCA}, we obtain for the Natural model 99.5\%, 98\%, 99,9\% and 100\% evasion rates on $dFGSM^s$, $rFGSM^s$, $BGA^s$ and $BCA^s$, respectively. 
We show that the four evasion attacks on the PCA DNNs architecture are significantly increase the evasion rate against the Natural training model.
Also, the overall accuracy increased for all five models from range 86-88\% in Experiment I as shown in Table \ref{metrics} to range 93-93.5\% in Experiment II as shown in Table \ref{metricsPCA}.
In experiment I, the evasion rate for $BCA^s$ is 64\% while the evasion rate in Experiment II for $BCA^s$ is 100\%. We observe that the values of CN for the $BCA^s$ in our two set of Experiments I and II have influenced the results. The CN ratios are 1.2 and 1.0 in experiment I and II as shown in Table \ref{metrics} and Table \ref{metricsPCA}, respectively. It is worth mentioning that the CN ratio for for the $BCA^s$ in \cite{al2018adversarial} was 0.

\begin{table}[h]
\begin{tiny}
\caption{Evasion Rate Results of Experiment II with PCA}
\label{evasionRatePCA}
\resizebox{\columnwidth}{!}{
\begin{tabular}{ |c|c|c|c|c| }
\hline
- & \multicolumn{4}{|c|}{Adversarial Attack Methods}\\
\hline
Model &  dFGSM & rFGSM & BGA & BCA\\
\hline
Natural &         \cellcolor{blue!25}\textbf{ 99.5} &     \cellcolor{blue!25}\textbf{98.0} &\cellcolor{blue!25}\textbf{99.9} & \cellcolor{blue!25}\textbf{100} \\
\hline
dFGSM &           4.1 &      4.1 &    4.1 &    4.1 \\
\hline
rFGSM &           3.8 &      3.8 &    3.8 &    3.8  \\
\hline
BGA  &      \cellcolor{yellow!25}{3.0} &    \cellcolor{yellow!25} {2.9} &  \cellcolor{yellow!25} {2.9} &   \cellcolor{yellow!25} {2.9} \\
\hline
BCA    &      5.0 &      4.9 &    4.6 &    4.6  \\
\hline
\end{tabular}
}
\end{tiny}
\end{table}

\begin{table}[h]
\begin{tiny}
\caption{Evaluation  Metrics  including  Overall  Accuracy, Evasion rate of models about the corresponding adversary (FNR), FPR and the Cover Number (CN)) rate for Experiment II}
\label{metricsPCA}
\resizebox{\columnwidth}{!}{

\begin{tabular}{ |c|c|c|c|c| } 
\hline
Model &  Accuracy &  FPR & Evasion Rate & CN ratio \\ 

\hline
Natural &    93.4 &  7.2 &  5.9 & 1\\
\hline
dFGSM &      93.7 &  8.1 &  4.1 & 0.8\\   
\hline
rFGSM &      93.9 &  8.0 &  3.8 & 1.1\\
\hline
BGA   &     93.0 & 10.4 &  2.9& 1\\
\hline
BCA   &      93.5 &  8.1 &  4.6&  1\\
\hline
\end{tabular}
}
\end{tiny}
\end{table}


To summarize the overall results of Experiment II, the evasion rates are relatively three-time lower as shown in Table \ref{evasionRatePCA} comparing to the evasion rates in Experiment I as shown in Table \ref{evasionRate}. We noticed that using PCA improves the resilience of IDS models against adversarial samples.
Moreover, the results in Experiment II Table \ref{evasionRatePCA} of the trained models show equal robustness against all adversaries in each method. For example, the result for $BGA^s$ with shade cells in yellow have the lowest and almost the same evasion rates 2.9-3\% for all adversarial attack methods. Also, $dFGSM^s$ model shows a higher resilience to adversarial attack comparing to $dFGSM^s$ model in Experiment I. The evasion rates reduce from 15.7-23\% in Experiment I to range of 2.9-3\% as shown in Table \ref{evasionRatePCA} in Experiment II.
Another observation is that the FNR and FPR values in Experiment II as shown in Table \ref{metricsPCA} in generally lower than the FNR in Experiment I as shown in Table \ref{metrics}. Indeed, FNR and FPR are reduced because the adversarial training models robustness improve. 

\subsection{Discussion}
We highlighted in our evaluation of our IDS prototype the following aspects:
Does our IDS prototype based on min-max (or saddle-point) formulation improve the robustness against adversarial samples generated by the inner maximization?
Does the PCA reduce the evasion rate of the studied adversarial attacks?
Do adversarial attack methods that are initially designed \cite{al2018adversarial} to attack deep learning models in discrete/binary features domain exhibit different misclassification levels in continuous feature domain?

The experiments show that the robustness of our Natural IDS model (defender) that is trained with a clean dataset has high evasion rate of 100\% against the four adversarial attacks introduced in previous sections. While the IDS models trained using min-max (or saddle-point) approach are more robust to adversarial samples as shown in Table \ref{evasionRate} and \ref{evasionRatePCA} among all adversaries attacks samples. By using the inner maximizer approach, we generate adversarial samples that maximize the evasion rate. 
We assess the robustness of our models that are trained using the  adversarial training methods and we achieve low evasion rates across all the adversarial attack methods.   
Likewise, experiment II shows that performing PCA on our selected dataset leads to decrease the overall evasion rates across all adversarial attack methods. 
In this paper, although we have continuous feature space, we used $BGA^s$ and $BCA^s$ in our work to generate adversarial samples and incorporate them in the training phase. It is interesting to note that $BGA^s$ and $BCA^s$ adversarial attack methods \cite{al2018adversarial} that have been designed to fit discrete data like binary malware file have the highest evasion rate among all other used adversarial attack methods. 

Using different adversarial attack methods with the saddle-point formulation in deep learning-based IDS in addition to malware detection \cite{al2018adversarial} and image classification \cite{madry2017towards} that used also saddle-point formulation, opens a new direction to explore general defense against more type of adversarial attacks.

\section{Conclusion and future work}
In this work, we have applied the min-max (or saddle-point) formulation in the IDS domain and investigated the effectiveness of the "inner maximization problem" on the robustness of adversarially trained model "outer minimization problem."
We generated adversarial samples using four existing methods $dFGSM^s$, $rFGSM^s$,$BGA^s$ and $BCA^s$. We analyzed if $BGA^s$ and $BCA^s$, which are designed for discrete feature domain can generate adversarial samples in continuous features domain. We found out that BGA adversaries have the highest evasion rate among all other adversaries. We also found that all other trained models are considerably vulnerable to the BGA and BCA adversaries. Our experiments provide evidence that DNN with min-max (or saddle-point) formulation increases the robustness of the experimental IDS using the UNSW-NB 15 dataset. Moreover, we used Principal Component Analysis (PCA), and the experiments showed that carrying out dimensionality reduction using PCA on the dataset helped in decreasing evasion rates. 
We believe that further exploration in dimensionality reduction in the deep neural networks can lead us to further optimization and robustness of IDS solutions.

\bibliographystyle{plain}
\bibliography{references}



\end{document}